\documentclass[conference]{IEEEtran}
\usepackage[utf8]{inputenc}
\usepackage{url}
\usepackage{amsmath}
\usepackage{amssymb}
\usepackage{graphicx}
\usepackage{float}
\usepackage{booktabs}
\usepackage{multirow}
\usepackage{cite}
\usepackage{algorithm}
\usepackage{algpseudocode}
\usepackage{hyperref}
\usepackage{fontawesome5}
\usepackage{tikz}
\usetikzlibrary{positioning, calc}
\definecolor{aggcolor}{RGB}{205,112,93}
\definecolor{opcolor}{RGB}{180,30,30}
\definecolor{hemscolor}{RGB}{0,128,128}
\definecolor{usercolor}{RGB}{0,114,178}

\title{Conversational Demand Response: Bidirectional\\ Aggregator-Prosumer Coordination through Agentic~AI}

\author{
\IEEEauthorblockN{Reda El Makroum\textsuperscript{1,*}, Sebastian Zwickl-Bernhard\textsuperscript{1,2}, Lukas Kranzl\textsuperscript{1}, Hans Auer\textsuperscript{1,2}}
\IEEEauthorblockA{\textsuperscript{1}Energy Economics Group, TU Wien, Gu{\ss}hausstra{\ss}e 25--29, Vienna, Austria}
\IEEEauthorblockA{\textsuperscript{2}Industrial Economics and Technology Management, NTNU, Trondheim, Norway}
\IEEEauthorblockA{\textsuperscript{*}elmakroum@eeg.tuwien.ac.at}
}

\begin{document}

\maketitle

\begin{abstract}
Residential demand response depends on sustained prosumer participation, yet existing coordination is either fully automated, or limited to one-way dispatch signals and price alerts that offer little possibility for informed decision-making. This paper introduces Conversational Demand Response (CDR), a coordination mechanism where aggregators and prosumers interact through bidirectional natural language, enabled through agentic AI. A two-tier multi-agent architecture is developed in which an aggregator agent dispatches flexibility requests and a prosumer Home Energy Management System (HEMS) assesses deliverability and cost-benefit by calling an optimization-based tool. CDR also enables prosumer-initiated upstream communication, where changes in preferences can reach the aggregator directly. Proof-of-concept evaluation shows that interactions complete in under 12 seconds. The architecture illustrates how agentic AI can bridge the aggregator-prosumer coordination gap, providing the scalability of automated DR while preserving the transparency, explainability, and user agency necessary for sustained prosumer participation. All system components, including agent prompts, orchestration logic, and simulation interfaces, are released as open source to enable reproducibility and further development.
\end{abstract}

\begin{IEEEkeywords}
Conversational demand response, agentic AI, home energy management systems, prosumer engagement, natural language coordination
\end{IEEEkeywords}

\section{Introduction}

Achieving net-zero targets requires global demand response (DR) capacity to reach 500 GW by 2030, with buildings and electric vehicles accounting for approximately 60\% of this potential \cite{iea_demand_2023}. Realizing this potential requires consistent prosumer participation, yet voluntary opt-in rates for residential DR programs rarely exceed 20\% \cite{fowlie_default_2021}. To understand why, Siitonen et al. \cite{siitonen_customer_2023} synthesize findings across European DR pilot projects and find that interaction quality, communication clarity, and easily interpretable information are critical determinants of sustained engagement. Parrish et al. \cite{parrish_systematic_2020} confirm these patterns across international trials, identifying complexity, perceived loss of control, and effort as primary barriers. As long as these barriers persist, the scheduling flexibility that residential loads offer \cite{su_exploring_2026} and that aggregators are designed to mobilize across market stages \cite{ottesen_multi_2018} remains largely unrealized.

At the aggregator-prosumer interface, coordination today is either fully automated or relies on static one-way notifications. Automating DR actions reduces cognitive load and increases willingness to participate \cite{valor_schemes_2025}, yet it undermines the transparency and user agency necessary to retain prosumers long-term \cite{diamond_encouraging_2023}. Fabianek et al. \cite{fabianek_multicriteria_2025} find that, beyond financial compensation, criteria such as guaranteed comfort, control, and transparency critically shape occupant acceptance of direct load control. What is needed, then, is a coordination mechanism that scales like automation but preserves the transparency and agency of direct interaction, a concept we term Conversational Demand Response (CDR). 

Recent advances in agentic AI, which harnesses large language models (LLMs) as autonomous tool-using agents capable of planning, reasoning, and interacting with external tools \cite{chowa_language_2025}, now make such coordination technically feasible. Energy applications have begun to exploit these capabilities across the DR value chain. On the household level, Michelon et al. \cite{michelon_large_2025} demonstrate that an LLM interface can translate informal user inputs into structured HEMS parameters through conversation, while others have explored cost-optimal scheduling via model predictive control \cite{raghavan_costoptimal_2025}, residential energy advisory \cite{gkalinikis_rhea_2025}, and context-aware device control in smart buildings \cite{he_contextaware_2025}. These systems, however, operate within a single household without coordinating with external market actors; they advise prosumers but do not coordinate with aggregators. On the aggregator side, multi-agent frameworks have achieved consumption reduction through coordinated DR \cite{rashed_aidriven_2025}, LLM-driven digital twins have predicted user behavior for incentive design \cite{sun_dynamic_2025}, and LLM-powered platforms have been proposed to enable prosumer participation in flexibility markets \cite{jatowt_flexidigital_2025}.

These two streams remain disconnected: prosumer-facing applications lack aggregator integration, while aggregator-side applications lack conversational prosumer engagement. The closest work to bridging these layers is Zhang et al. \cite{zhang_twolayer_2025}, who deploy LLMs within a two-layer aggregator-EV framework. Each EV user is modeled as an LLM-driven agent that generates charging decisions from individualized profiles and dynamic price signals, while the aggregator optimizes retail prices to maximize profit. However, the LLMs simulate user behavior rather than enabling conversational interaction with prosumers. To our knowledge, no existing work addresses this coordination gap.

This gap motivates the following research question: How can agentic AI realize CDR that is both scalable for aggregators and transparent for prosumers? We propose a two-tier multi-agent system, building on our prior agentic HEMS architecture \cite{elmakroum_agentic_2025}, where both aggregator and prosumer operate as LLM-based agentic systems coordinating through natural language, with the HEMS leveraging optimization-based sub-agents to ground conversations in quantified feasibility assessments. We present three contributions:

\begin{itemize}\setlength{\itemsep}{0pt}\setlength{\parskip}{0pt}
    \item We design a two-tier multi-agent architecture for CDR, where an aggregator agent and a prosumer HEMS agent coordinate through bidirectional natural language, replacing one-way dispatch with transparent, interactive exchanges.
    \item We ground these conversations in quantified feasibility assessments through asset-level sub-agents that can call a MILP optimizer as a tool, enabling the HEMS to present cost-benefit trade-offs to the prosumer in natural language before any commitment is made.
    \item We demonstrate end-to-end CDR operation through downstream and upstream proof-of-concept scenarios, showing that the two-tier system can coordinate, evaluate, and execute DR events with the prosumer in the loop. The full implementation is released as open source\footnote{All system components, including agent prompts, orchestration logic, optimizer code, and simulation interfaces, are available at: \url{https://github.com/RedaElMakroum/cdr}} to enable replication and further CDR research.
\end{itemize}
\section{Conversational Demand Response}

We introduce Conversational Demand Response (CDR) as a coordination mechanism between aggregators and prosumers, where agentic AI systems provide decision support on both sides of the interaction.

Current DR coordination faces three structural limitations. First, existing systems lack \textit{transparency and explainability}: prosumers receive dispatch signals without understanding what is requested or whether compensation reflects their actual flexibility value. Second, coordination lacks \textit{real-time adaptability}: aggregators cannot adjust requests to a prosumer's current situation, and prosumers cannot communicate changing constraints before activation. Third, interaction remains \textit{unidirectional}: aggregators dispatch and prosumers comply or opt out, with no mechanism to coordinate terms or proactively offer flexibility. These limitations compound the behavioral barriers that erode sustained DR participation \cite{sloot_behavioral_2023}.

CDR addresses these limitations through bidirectional conversational interaction. Because the exchange is text-based and mediated by LLMs, every technical result, from optimizer outputs to asset constraints, can be translated into plain-language explanations accessible to non-expert prosumers. Downstream, the aggregator's agentic system identifies flexibility opportunities from market signals, formulates DR events with context and compensation, and dispatches requests to prosumer agents. The prosumer's HEMS evaluates feasibility per load, presents options in natural language, and executes only upon approval. Upstream, the prosumer's HEMS maintains a persistent connection to the aggregator, enabling schedule changes, preferences, or constraints to propagate directly.

\begin{figure}[t]
    \centering
    \includegraphics[width=\columnwidth]{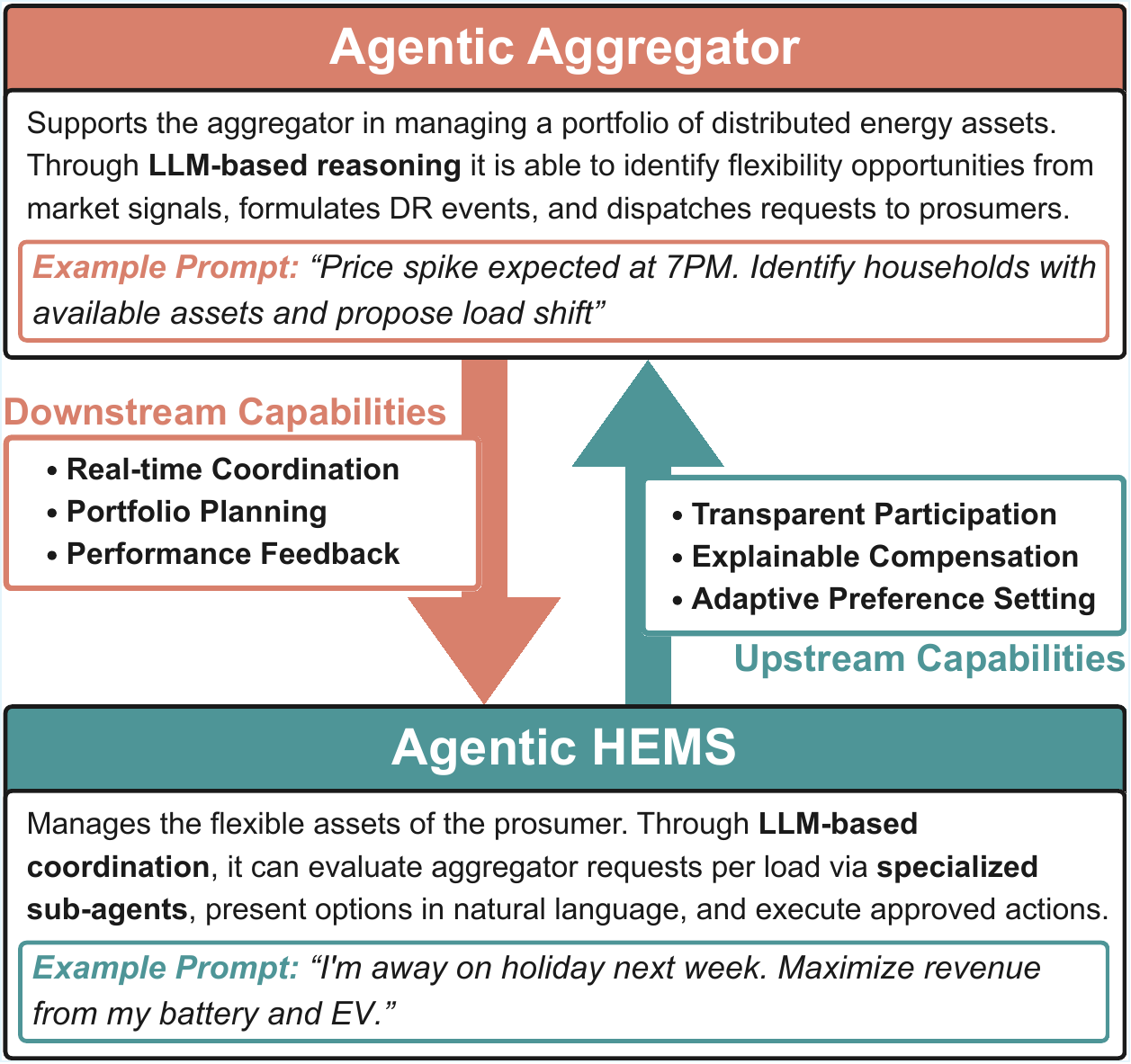}
    \caption{Bidirectional communication in CDR.}
    \label{fig:cdr_overview}
    \vspace{-1.5em}
\end{figure}

Consider a prosumer who, before a week-long absence, instructs their HEMS to maximize revenue from available assets. The system translates this into a flexibility offer and communicates it to the aggregator, who updates portfolio planning accordingly. Fig.~\ref{fig:cdr_overview} presents the CDR interaction framework.

Each capability addresses one or more of these limitations. In practice, a prosumer who just plugged in their EV receives a request that accounts for their current battery state, not a generic peak-shaving command. Portfolio planning gives prosumers visibility into upcoming grid needs, allowing them to prepare rather than react. Performance feedback quantifies what each household contributed and earned, building the trust that sustains repeated participation. Table~\ref{tab:cdr_downstream} summarizes the three downstream capabilities.

\begin{table}[!ht]
\vspace{-1em}
\centering
\caption{Downstream Capabilities Enabled by CDR}
\label{tab:cdr_downstream}
\footnotesize
\begin{tabular*}{\columnwidth}{@{}l@{\extracolsep{\fill}}p{4.9cm}@{}}
\toprule
\textbf{Capability} & \textbf{Description} \\
\midrule
Real-time Coordination & Aggregator adapts requests to prosumer context rather than issuing static commands \\
\midrule
Portfolio Planning & Aggregator communicates upcoming events, enabling prosumers to plan ahead \\
\midrule
Performance Feedback & Prosumers receive quantified impact of past participation, reinforcing engagement \\
\bottomrule
\end{tabular*}
\end{table}
A prosumer asked to reduce consumption at 18:00 sees exactly why the request was made, which loads are affected, and what they stand to earn. Compensation links each reward to the specific market conditions that generated it, replacing flat-rate incentives with verifiable value. When a household buys an EV or installs a heat pump, it can update its flexibility profile without re-enrolling in the DR program. Table~\ref{tab:cdr_upstream} summarizes the upstream capabilities. 

\begin{table}[!ht]
\centering
\caption{Upstream Capabilities Enabled by CDR}
\label{tab:cdr_upstream}
\footnotesize
\begin{tabular*}{\columnwidth}{@{}l@{\extracolsep{\fill}}p{4.9cm}@{}}
\toprule
\textbf{Capability} & \textbf{Description} \\
\midrule
Transparent Participation & Prosumers see the full impact and compensation in plain language before committing \\
\midrule
Explainable Compensation & Rewards are explained per load and linked to market conditions \\
\midrule
Adaptive Preference Setting & Prosumers can update constraints and preferences at any time, not only at enrollment \\
\bottomrule
\vspace{-3em}
\end{tabular*}
\end{table}
\newpage
The following section describes the multi-agent architecture that realizes these capabilities.

\section{System Architecture \& Implementation}

To enable conversational coordination, we implement CDR as a two-tier multi-agent system where both the aggregator and prosumer operate as LLM-based agentic systems coordinating through natural language. The implementation extends our prior agentic HEMS \cite{elmakroum_agentic_2025} with two new components: an aggregator agent that manages portfolio-level dispatch and upstream routing, and a battery HEMS sub-agent that uses a MILP optimizer as a tool to assess DR feasibility in real time. While appliance sub-agents handle discrete loads through simple scheduling, the battery requires joint optimization of charge, discharge, and grid exchange across all timeslots, warranting a dedicated MILP formulation that the sub-agent calls as a tool. The aggregator agent and the HEMS orchestrator follow the ReAct pattern \cite{yao_react_2023}, iterating between reasoning and action steps until a task is resolved. The aggregator agent receives market signals and dispatches contextualized DR events to registered households, while the HEMS orchestrator delegates evaluation to specialized sub-agents per load and translates results into conversational options for the prosumer.

\subsection{Aggregator Agent}

The aggregator agent coordinates portfolio-level dispatch by managing registered households and their distributed energy resources. The agent can receive market obligations or operator instructions, identify target households from registered capacity, and formulate DR events specifying the time window, target power, and compensation rate. The agent does not estimate household-level feasibility; it formulates the request and delegates assessment entirely to the prosumer's HEMS. This separation is deliberate. The aggregator operates on portfolio-level signals, while feasibility evaluation stays local to the household, where the HEMS has direct access to asset state and prosumer context. The aggregator also handles prosumer-initiated upstream messages. It classifies the message type and routes it accordingly. The agent applies asset updates and preference changes directly to the portfolio, and escalates contract modifications or complaints.

\subsection{HEMS Agent}

The prosumer-side HEMS receives and evaluates DR events dispatched by the aggregator. It operates as a hierarchical multi-agent system where an LLM-based orchestrator delegates to specialized sub-agents per household load. Each sub-agent manages a specific energy asset, evaluating scheduling constraints and operating conditions for its assigned load. Unlike the orchestrator, which coordinates through iterative ReAct cycles, sub-agents operate in single-turn interactions: they receive tailored inputs from the orchestrator and return a structured recommendation. This separation keeps the orchestrator focused on coordination while sub-agents handle domain-specific assessment. 

The orchestrator operates in two modes. In normal scheduling mode, it delegates to appliance agents and executes approved schedules. In DR event mode, activated by an incoming aggregator request, the orchestrator parses the request, reasons about which energy asset is best suited to respond, and delegates to the corresponding sub-agent for feasibility assessment. The sub-agent's technical result is then translated into a conversational explanation for the prosumer, covering the request details, deliverable capacity, expected impact, and compensation.

\subsection{Battery Optimizer}

The battery sub-agent uses a MILP optimizer as a callable tool. When the orchestrator delegates a DR feasibility query, the sub-agent formulates the problem and calls the optimizer, which determines the cost-optimal charge and discharge schedule over a 24-hour horizon at 15-minute resolution, minimizing net electricity cost, battery degradation, and peak discharge:
\begin{equation}
\begin{aligned}
\min \sum_{t} \Big[ & \left( p_t^{\text{imp}} \cdot \pi_t - p_t^{\text{exp}} \cdot \pi^{\text{fit}} \right) \Delta t \\
& + \left( p_t^{\text{ch}} + p_t^{\text{dis}} \right) c^{\text{deg}} \Delta t \Big] \\
& + d^{\text{peak}} \cdot w^{\text{peak}}
\end{aligned}
\label{eq:objective}
\end{equation}
where $p_t^{\text{imp}}$ and $p_t^{\text{exp}}$ denote grid import and export power, $\pi_t$ the day-ahead price, $\pi^{\text{fit}}$ the feed-in tariff (constant, reflecting fixed contractual rates typical in European prosumer agreements), $p_t^{\text{ch}}$ and $p_t^{\text{dis}}$ the battery charge and discharge power, $c^{\text{deg}}$ a degradation cost, $d^{\text{peak}}$ the peak discharge power, $w^{\text{peak}}$ a regularization weight, and $\Delta t = 0.25$~h. The objective is minimized subject to:
\begin{align}
p_t^{\text{imp}} + p_t^{\text{PV}} + p_t^{\text{dis}} &= d_t + p_t^{\text{exp}} + p_t^{\text{ch}} & &\forall t \label{eq:balance} \\
E_{t+1} &= E_t + \eta_{\text{ch}}\, p_t^{\text{ch}}\, \Delta t - \frac{p_t^{\text{dis}}}{\eta_{\text{dis}}}\, \Delta t & &\forall t \label{eq:soc} \\
E^{\min} &\leq E_t \leq E^{\max} & &\forall t \label{eq:soc_bounds} \\
p_t^{\text{dis}} &\leq d_t & &\forall t \label{eq:discharge_cap} \\
d^{\text{peak}} &\geq p_t^{\text{dis}} & &\forall t \label{eq:peak}
\end{align}
where $p_t^{\text{PV}}$ is PV generation, $d_t$ household demand, $E_t$ battery state of charge, and $\eta_{\text{ch}} = \eta_{\text{dis}} = \sqrt{\eta_{\text{RT}}}$ one-way efficiency losses. Constraint~\eqref{eq:balance} enforces energy balance per timeslot. Constraints~\eqref{eq:soc}--\eqref{eq:soc_bounds} track SoC dynamics and enforce capacity limits. Constraint~\eqref{eq:discharge_cap} caps discharge at demand, preventing battery-to-grid export outside DR events. Constraint~\eqref{eq:peak} tracks peak discharge for regularization. Charge/discharge and import/export are mutually exclusive via Big-M constraints. A self-consumption priority constraint (omitted for brevity) ensures PV surplus charges the battery before grid export.

For DR feasibility assessment, the optimizer performs a dual-solve procedure. It first solves the baseline without DR obligations, then re-solves with an additional constraint:
\begin{equation}
p_t^{\text{dis}} \geq p^{\text{DR}} \quad \forall\, t \in \mathcal{T}^{\text{DR}}
\label{eq:dr_constraint}
\end{equation}
forcing discharge above the aggregator's target during the requested window $\mathcal{T}^{\text{DR}}$. If the DR-committed problem is infeasible at the full requested power, a binary search identifies the maximum deliverable level. For same-day requests, the optimizer locks slots prior to request arrival to the baseline solution, since past dispatch actions are irreversible. The result includes a feasibility verdict (full, partial, or infeasible), the economics of participation, and the SoC trajectory under DR commitment.

\begin{figure}[h]
    \centering
    \includegraphics[width=\columnwidth]{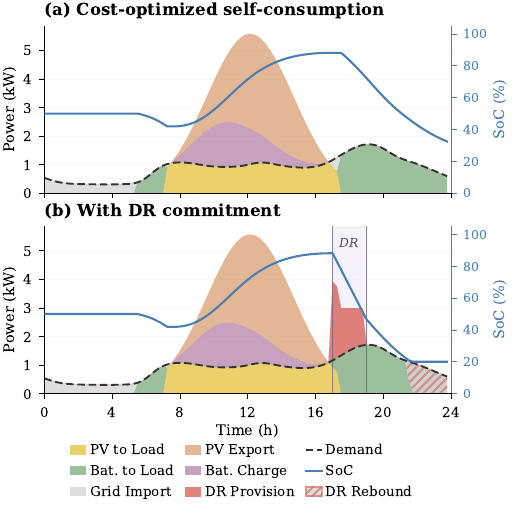}
    \caption{Illustration of the dual-solve procedure for a representative household day. (a)~Cost-optimized self-consumption baseline. (b)~Schedule with DR commitment.}
    \label{fig:load_profile}
\end{figure}

Fig.~\ref{fig:load_profile} illustrates this dual-solve procedure for a representative household day.

Panel~(a) shows the cost-optimized self-consumption schedule without a DR event: PV generation charges the battery during midday, and the battery discharges in the evening to cover household demand. Panel~(b) shows the same day with a DR commitment requesting discharge between 17:00 and 19:00. The optimizer pre-charges the battery to a higher SoC in anticipation and forces discharge during the requested window. The resulting rebound, visible as increased grid import after the event, represents the cost of participation. The optimizer weighs this additional import cost against the DR compensation and determines that participating is more economical than the self-consumption scenario. This cost-benefit comparison is what the HEMS translates into natural language and presents to the prosumer before any commitment is made.

\section{Results}

We demonstrate CDR through two end-to-end exchanges that exercise both directions of coordination: an aggregator-initiated DR dispatch and a prosumer-initiated portfolio update.

\subsection{Experimental Setup}

All agents leverage GPT-OSS-120B \cite{openai_gptoss_2025} with reasoning enabled and temperature set to zero to minimize output variability, served through the Cerebras inference API at up to 2,500 tokens per second \cite{he_waferllm_2025}. The HEMS battery optimization-based tool solves the MILP formulation using PuLP. Each scenario runs in an isolated agent session with independent context windows, ensuring no information leakage between the aggregator and HEMS sides. Agent responses are trimmed for space. Appendix~\ref{app:params} summarizes the simulation parameters.

\subsection{Downstream: Aggregator-Initiated DR Dispatch}
The operator requests 3\,kW of flexibility for the 17:00--19:00 evening window. The aggregator agent receives this request and dispatches it to the household.

\smallskip
\noindent\begin{tikzpicture}
\pgfmathsetlengthmacro{\W}{\columnwidth}
\pgfmathsetlengthmacro{\bw}{2.5pt}
\pgfmathsetlengthmacro{\pad}{0.3cm}
\pgfmathsetlengthmacro{\bub}{0.78\columnwidth}

\node[rectangle, rounded corners=3pt, inner sep=5pt, font=\scriptsize, text width=\bub, draw=opcolor, fill=white, anchor=north east] at (\W-\pad, 0) (m1) {We need 3\,kW of flexibility from the portfolio between 17:00 and 19:00. What can households offer?};
\node[rounded corners=3pt, fill=opcolor, inner sep=3.5pt, font=\scriptsize\sffamily\bfseries, text=white, anchor=south east] at ([yshift=3pt]m1.north east) {\faUser~Aggregator};

\node[rectangle, rounded corners=3pt, inner sep=5pt, font=\scriptsize, text width=\bub, draw=aggcolor, fill=white, anchor=north west] at (\pad, {-1.6}) (m2) {Dispatched a request to HH-001. Status: pending. I will monitor and report the household's decision as soon as it is available.};
\node[rounded corners=3pt, fill=aggcolor, inner sep=3.5pt, font=\scriptsize\sffamily\bfseries, text=white, anchor=south west] at ([yshift=3pt]m2.north west) {\faRobot~Agentic Aggregator};

\fill[aggcolor] (0,0.7) rectangle (\bw,-2.75);
\fill[opcolor] (\W,0.7) rectangle (\W-\bw,-2.75);

\end{tikzpicture}

\noindent The agentic aggregator autonomously queries its portfolio, identifies HH-001 as most suitable, and dispatches the event. The agentic HEMS then receives the request.

\smallskip
\noindent\begin{tikzpicture}
\pgfmathsetlengthmacro{\W}{\columnwidth}
\pgfmathsetlengthmacro{\bw}{2.5pt}
\pgfmathsetlengthmacro{\pad}{0.3cm}
\pgfmathsetlengthmacro{\bub}{0.78\columnwidth}

\node[rectangle, rounded corners=3pt, inner sep=4pt, font=\scriptsize, text width=0.6\columnwidth, draw=hemscolor, fill=white, anchor=north west] at (\pad, 0) (m1) {DR event received. Delegated to battery agent. Fully feasible. Net benefit: \texteuro\,1.13. SoC: 30\,\% $\rightarrow$ 43\,\%. No comfort impact. Would you like me to accept?};
\node[rounded corners=3pt, fill=hemscolor, inner sep=3.5pt, font=\scriptsize\sffamily\bfseries, text=white, anchor=south west] at ([yshift=2pt]m1.north west) {\faRobot~Agentic HEMS};

\node[rectangle, rounded corners=3pt, inner sep=4pt, font=\scriptsize, text width=0.25\columnwidth, draw=usercolor, fill=white, anchor=north east] at (\W-\pad, {-1.5}) (m2) {Yes, go ahead.};
\node[rounded corners=3pt, fill=usercolor, inner sep=3.5pt, font=\scriptsize\sffamily\bfseries, text=white, anchor=south east] at ([yshift=2pt]m2.north east) {\faUser~Prosumer};

\node[rectangle, rounded corners=3pt, inner sep=4pt, font=\scriptsize, text width=0.6\columnwidth, draw=hemscolor, fill=white, anchor=north west] at (\pad, {-2.4}) (m3) {Done. Commitment submitted to aggregator.};
\node[rounded corners=3pt, fill=hemscolor, inner sep=3.5pt, font=\scriptsize\sffamily\bfseries, text=white, anchor=south west] at ([yshift=2pt]m3.north west) {\faRobot~Agentic HEMS};

\fill[hemscolor] (0,0.7) rectangle (\bw,-3.1);
\fill[usercolor] (\W,0.7) rectangle (\W-\bw,-3.1);

\end{tikzpicture}

The HEMS reasons through the request, and decides to delegate to the battery sub-agent, which in turn calls the optimization-based tool. The dual-solve confirms full feasibility without affecting comfort. The HEMS translates the optimizer output into the conversational summary shown above. Upon approval, the HEMS submits the commitment and the aggregator closes the loop.

\smallskip
\noindent\begin{tikzpicture}
\pgfmathsetlengthmacro{\W}{\columnwidth}
\pgfmathsetlengthmacro{\bw}{2.5pt}
\pgfmathsetlengthmacro{\pad}{0.3cm}
\pgfmathsetlengthmacro{\bub}{0.78\columnwidth}

\node[rectangle, rounded corners=3pt, inner sep=4pt, font=\scriptsize, text width=0.6\columnwidth, draw=aggcolor, fill=white, anchor=north west] at (\pad, 0) (m1) {HH-001 accepted the flexibility request. Commitment confirmed, portfolio updated.};
\node[rounded corners=3pt, fill=aggcolor, inner sep=3.5pt, font=\scriptsize\sffamily\bfseries, text=white, anchor=south west] at ([yshift=2pt]m1.north west) {\faRobot~Agentic Aggregator};

\fill[aggcolor] (0,0.7) rectangle (\bw,-1.0);
\fill[opcolor] (\W,0.7) rectangle (\W-\bw,-1.0);
\end{tikzpicture}

\noindent This concludes the downstream exchange simulation. Because the interaction is conversational, the prosumer can ask follow-up questions, request alternative schedules, or reject the event at any point, ensuring transparency, explainability, and user agency throughout.

\medskip
\subsection{Upstream: Prosumer-Initiated Profile Update}
CDR also supports prosumer-initiated communication, where for instance a prosumer leaving for holiday notifies the HEMS, which propagates the updated availability to the aggregator.

\smallskip
\noindent\begin{tikzpicture}
\pgfmathsetlengthmacro{\W}{\columnwidth}
\pgfmathsetlengthmacro{\bw}{2.5pt}
\pgfmathsetlengthmacro{\pad}{0.3cm}
\pgfmathsetlengthmacro{\bub}{0.78\columnwidth}

\node[rectangle, rounded corners=3pt, inner sep=4pt, font=\scriptsize, text width=0.6\columnwidth, draw=usercolor, fill=white, anchor=north east] at (\W-\pad, 0) (m1) {I'm away on holiday next week. Maximize revenue from my battery and EV.};
\node[rounded corners=3pt, fill=usercolor, inner sep=3.5pt, font=\scriptsize\sffamily\bfseries, text=white, anchor=south east] at ([yshift=2pt]m1.north east) {\faUser~Prosumer};

\node[rectangle, rounded corners=3pt, inner sep=4pt, font=\scriptsize, text width=0.6\columnwidth, draw=hemscolor, fill=white, anchor=north west] at (\pad, {-1.3}) (m2) {Noted. Switching to revenue maximization for next week. Aggregator notified.};
\node[rounded corners=3pt, fill=hemscolor, inner sep=3.5pt, font=\scriptsize\sffamily\bfseries, text=white, anchor=south west] at ([yshift=2pt]m2.north west) {\faRobot~Agentic HEMS};

\fill[hemscolor] (0,0.7) rectangle (\bw,-2.1);
\fill[usercolor] (\W,0.7) rectangle (\W-\bw,-2.1);

\end{tikzpicture}

\noindent The HEMS acknowledges the change and informs the aggregator, which updates its portfolio accordingly.

\smallskip
\noindent\begin{tikzpicture}
\pgfmathsetlengthmacro{\W}{\columnwidth}
\pgfmathsetlengthmacro{\bw}{2.5pt}
\pgfmathsetlengthmacro{\pad}{0.3cm}
\pgfmathsetlengthmacro{\bub}{0.78\columnwidth}

\node[rectangle, rounded corners=3pt, inner sep=4pt, font=\scriptsize, text width=0.78\columnwidth, draw=aggcolor, fill=white, anchor=north west] at (\pad, 0) (m1) {HH-001 will be away next week and has granted full asset access for DR dispatch. I've re-optimized the portfolio accordingly.};
\node[rounded corners=3pt, fill=aggcolor, inner sep=3.5pt, font=\scriptsize\sffamily\bfseries, text=white, anchor=south west] at ([yshift=2pt]m1.north west) {\faRobot~Agentic Aggregator};

\fill[aggcolor] (0,0.7) rectangle (\bw,-1.0);
\fill[opcolor] (\W,0.7) rectangle (\W-\bw,-1.0);

\end{tikzpicture}
\vspace{-1.5em}
\smallskip

Together, the downstream and upstream scenarios illustrate the full CDR loop: an aggregator can dispatch a contextualized flexibility request and obtain a prosumer commitment backed by optimization-based feasibility assessment, while the prosumer retains the ability to understand, question, and initiate changes at any point. 

\subsection{Benchmarking}
For CDR to be practical, the latency between messages should remain within conversational expectations and the token cost per interaction (tokens are the units of text processed by the LLM, determining inference cost) needs to be reasonable for operational deployment. To assess this, we benchmark six scenarios that span the range of interactions CDR must handle. The three downstream scenarios cover increasing coordination complexity: a standard acceptance, a rejection, and a high-target request (5\,kW) where the business-as-usual charging schedule alone cannot meet the target and the optimizer must plan pre-charging from forecasted PV. The three upstream scenarios cover the main profile update types: an availability change, a preference modification, and a new asset registration. Each scenario is executed five times to capture run-to-run variability in LLM inference (Table~\ref{tab:scenarios}).

\begin{table}[h]
\vspace{-1em}
\centering
\caption{CDR Computational Feasibility (mean $\pm$ std, $n{=}5$)}
\label{tab:scenarios}
\scriptsize
\begin{tabular*}{\columnwidth}{@{\extracolsep{\fill}}llcccc@{}}
\toprule
Direction & Scenario & Iterations & Tool calls & Tokens (k) & Time (s) \\
\midrule
\multirow{3}{*}{Downstream} & Acceptance      & 3.6$\pm$0.5  & 2.6$\pm$0.5  & 23.4$\pm$4.0  & 8.3$\pm$2.1  \\
                             & Rejection       & 5.0$\pm$0.7  & 3.6$\pm$0.5  & 34.2$\pm$6.0  & 9.8$\pm$1.9  \\
                             & High-target     & 3.4$\pm$0.5  & 2.4$\pm$0.5  & 21.8$\pm$4.1  & 7.8$\pm$2.2  \\
\midrule
\multirow{3}{*}{Upstream}   & Availability    & 1  & 0  & 1.3$\pm$0.0  & 1.3$\pm$0.8  \\
                             & Preference      & 1  & 0  & 1.0$\pm$0.0  & 1.7$\pm$1.6  \\
                             & New Asset       & 1  & 0  & 1.6$\pm$0.1  & 1.3$\pm$1.2  \\
\bottomrule
\end{tabular*}
\vspace{-1em}
\end{table}

Downstream scenarios require 3--5 reasoning iterations and 2--4 tool calls, completing in 7.8--9.8\,s on average. The rejection scenario incurs the highest cost (5.0$\pm$0.7 iterations, 34.2k tokens) because the agent still executes the full feasibility workflow (battery state retrieval, optimizer call, sub-agent evaluation), then additionally composes a detailed rejection explanation incorporating the prosumer's reasoning before relaying it to the aggregator. Acceptance follows a shorter path: feasibility check, prosumer approval, submission. The high-target scenario confirms that the battery sub-agent correctly adapts the optimizer parameters to a higher power target and still achieves full feasibility through PV pre-charging. Upstream scenarios resolve in a single iteration without tool calls (1.1--1.7\,s), as they require only message classification and profile routing rather than optimization. Across all six scenarios, no interaction exceeds 12 seconds or 35,000 tokens, and the agents produce consistent outputs across repeated runs, indicating that the architecture can sustain conversational interaction speeds necessary for real-time transparency and prosumer engagement.

\section{Discussion and Conclusion}

The results reveal that current LLM capabilities are sufficient to enable bidirectional, natural-language DR coordination between aggregators and prosumers. Both directions complete within seconds, though latency under concurrent multi-household load remains to be validated.
In practice, CDR is not intended to replace automated control for routine operations. Automated scheduling continues to handle execution; CDR adds a conversational layer that engages prosumers in their participation and makes the system legible to non-technical users. For this to reach households at scale, utilities and aggregation services would need to integrate such interfaces into their platforms. Our setup relies on cloud-based inference at up to 2,500 tokens per second, but advances in edge AI and model quantization are enabling local deployment on private hardware \cite{dritsas_deployment_2026}, which can eliminate cloud latency while preserving data privacy.

This work developed CDR as a coordination mechanism that replaces one-way dispatch with bidirectional natural-language interaction between aggregators and prosumers, and provided a proof of concept demonstrating its feasibility within operational DR timescales. The full implementation is released as open source to enable future extensions of the architecture with additional assets, optimization tools, and sub-agents on both the household and aggregator side. Field trials comparing conversational and conventional DR interfaces are needed to validate CDR's impact on prosumer engagement, and extending the framework to multi-household portfolio coordination remains a natural next step. Beyond the aggregator-prosumer level, the conversational coordination pattern may apply to higher grid layers, such as DSO-TSO interaction, where similar challenges of transparency and multi-actor coordination persist.
\clearpage
\bibliographystyle{IEEEtran}
\bibliography{references}

\appendix
\section{Simulation Parameters}
\label{app:params}

Table~\ref{tab:household} lists the battery, economic, and general parameters used in the proof-of-concept simulation. The open-source implementation, including agent prompts, orchestration logic, and simulation interfaces, is available at: \url{https://github.com/RedaElMakroum/cdr}.

\begin{table}[h]
\centering
\caption{Simulation Parameters}
\label{tab:household}
\footnotesize
\begin{tabular}{ll}
\toprule
\textbf{Parameter} & \textbf{Value} \\
\midrule
\multicolumn{2}{l}{\textit{Battery}} \\
Capacity & 15 kWh \\
Charge/discharge rate & 8 kW \\
Round-trip efficiency & 92\% \\
SoC bounds & 20--90\% \\
Initial SoC & 30\% (4.5 kWh) \\
\midrule
\multicolumn{2}{l}{\textit{Economic}} \\
Feed-in tariff & 0.04 EUR/kWh \\
Degradation cost & 0.015 EUR/kWh \\
DR compensation & 0.20 EUR/kWh \\
\midrule
\multicolumn{2}{l}{\textit{General}} \\
PV forecast & 8.5 kWh/day \\
Time resolution & 15 min (96 slots/day) \\
\bottomrule
\end{tabular}
\end{table}

\end{document}